\documentclass{article}

\usepackage{arxiv}
\usepackage[utf8]{inputenc} 
\usepackage[T1]{fontenc}    
\usepackage{hyperref}       
\usepackage{url}            
\usepackage{amsfonts}       
\usepackage{nicefrac}       
\usepackage{microtype}      
\usepackage{lipsum}
\usepackage{amsmath}
\usepackage{tabularx,multirow,subcaption,graphicx}
\usepackage{soul, color}
\usepackage{cleveref}

\def\v#1{\mathbf{#1}}
\def\m#1{\mathsf{#1}}
\def\s#1{\mathcal{#1}}

\def\tr#1{{#1}^\mathsf{T}}

\usepackage{authblk}
\author[1]{Sophia Bano}
\author[2,3]{Alessandro Casella}
\author[1]{Francisco Vasconcelos}
\author[4]{Sara Moccia}
\author[5,8]{George Attilakos}
\author[5,8]{Ruwan Wimalasundera}
\author[5,7,8]{Anna L. David}
\author[6]{Dario Paladini}
\author[7,8]{Jan Deprest}
\author[3]{Elena De Momi}
\author[2]{Leonardo S. Mattos}
\author[1]{Danail Stoyanov}

\affil[1]{Wellcome/EPSRC Centre for Interventional and Surgical Sciences(WEISS) and Department of Computer Science, University College London, London, UK}
\affil[2]{Department of Advanced Robotics, Istituto Italiano di Tecnologia, Genoa, Italy}
\affil[3]{Department of Electronics, Information and Bioengineering, Politecnico di Milano, Milan, Italy}
\affil[4]{The BioRobotics Institute, Scuola Superiore Sant’Anna, Pisa, Italy}
\affil[5]{Fetal Medicine Unit, Elizabeth Garrett Anderson Wing, University College London Hospital, London, UK}
\affil[6]{Department of Fetal and Perinatal Medicine, Istituto “Giannina Gaslini”, Genoa, Italy}
\affil[7]{Department of Development and Regeneration, University Hospital Leuven, Leuven, Belgium}
\affil[8]{EGA Institute for Women’s Health, Faculty of Population Health Sciences, University College London, UK}

{
    \makeatletter
    \renewcommand\AB@affilsepx{: \protect\Affilfont}
    \makeatother

    \affil[ ]{Correspondance}

    \makeatletter
    \renewcommand\AB@affilsepx{, \protect\Affilfont}
    \makeatother

    \affil[1]{sophia.bano@ucl.ac.uk}

}

\title{FetReg: Placental Vessel Segmentation and Registration in Fetoscopy Challenge Dataset}

\begin{document}
\maketitle

\begin{abstract}
Fetoscopy laser photocoagulation is a widely used procedure for the treatment of Twin-to-Twin Transfusion Syndrome (TTTS), that occur in mono-chorionic multiple pregnancies due to placental vascular anastomoses. This procedure is particularly challenging due to limited field of view, poor manoeuvrability of the fetoscope, poor visibility due to fluid turbidity, variability in light source, and unusual position of the placenta. This may lead to increased procedural time and incomplete ablation, resulting in persistent TTTS. Computer-assisted intervention may help overcome these challenges by expanding the fetoscopic field of view through video mosaicking and providing better visualization of the vessel network. However, the research and development in this domain remain limited due to unavailability of high-quality data to encode the intra- and inter-procedure variability. Through the \textit{Fetoscopic Placental Vessel Segmentation and Registration (FetReg)} challenge, we present a large-scale multi-centre dataset for the development of generalized and robust semantic segmentation and video mosaicking algorithms for the fetal environment with a focus on creating drift-free mosaics from long duration fetoscopy videos. In this paper, we provide an overview of the FetReg dataset, challenge tasks, evaluation metrics and baseline methods for both segmentation and registration. Baseline methods results on the FetReg dataset shows that our dataset poses interesting challenges, offering large opportunity for the creation of novel methods and models through a community effort initiative guided by the FetReg challenge. 
\end{abstract}


\section{Introduction}
Twin-to-twin Transfusion Syndrome~(TTTS) is a rare complication of monochorionic twin pregnancies, affecting 10-15\% of monochorionic diamniotic pregnancies. TTTS is characterized by the development of an unbalanced and chronic blood transfer from one twin (the donor twin), to the other (the recipient twin), through placental anastomoses~\cite{Baschat2011}.
This shared circulation causes profound fetal hemodynamic unbalance and consequently severe growth restriction, cardiovascular dysfunction, profound anaemia, hypovolaemic shock, circulatory collapse, hypoxic brain damage and death of one or both twins~\cite{Lewi2013}.

In 2004, a randomized, controlled trial demonstrated that fetoscopic laser ablation of placental anastomoses in TTTS resulted in a higher survival rate for at least one twin compared with other treatments, such as serial amnioreduction. Laser ablation further showed lower incidence of complications, such as cystic periventricular leukomalacia and neurologic complications~\cite{Senat2004,Baud2013}. 
The reported mortality for untreated TTTS was 90\%~\cite{pmid19445374}, after the implementation of laser therapy overall survival is around 63\%, and survival of at least one twin can reach 76\% of pregnancies~\cite{Muratore2009}.
A comprehensive description of all the steps that have led to consider laser surgery for coagulation of placental anastomoses the treatment of choice for TTTS can be found in~\cite{Deprest2010}.

Two types of laser ablation were studied in the past. The first (i) method is to laser coagulate all vessels that appear on fetoscopic examination to be an anastomosis. This is a non-reproducible and operator-dependent technique, which preserves placental function but which can miss important anastomoses. The second (ii) method, which reduces placental damage over (i), is laser coagulation of all vessels emerging from the amniotic membrane insertion, which relies on the assumption that all of these are vascular anastomoses~\cite{Quintero2007}. 
Today, the recognized elective treatment is the selective laser photocoagulation of communicating vessels originating in the donor’s placental territory. This method requires precise identification and laser ablation of placental vascular anastomoses. It relies on the classification of anastomoses in arterio-venous (from donor to recipient, AVDR, or from recipient to donor, AVRD), arterio-arterial (AA) or veno-venous (VV), and may distribute themselves following a certain pattern on the placenta. Recent researches have identified that the sequence in which these anastomoses are treated with the laser could result in further hypotension of the donor twin, with increased risks of complications and fetal demise~\cite{Nakata2009}. 

However, significant complication or failures were recorded in both techniques, the reason being that tiny anastomoses might be overlooked because flattened on the placental surface by the high pressure in the recipient’s sac. Therefore, the Solomon technique was introduced~\cite{Slaghekke2016} and now has become the gold standard in those cases in which it can be technically achieved. This technique consists in drawing a coagulation line connecting all the sites in which anastomoses had been coagulated after the first selective ablation round.

Despite all the advancements in instrumentation and imaging for TTTS~\cite{Cincotta2016}, residual anastomoses still represent a major complication in monochorionic placentas treated with fetoscopic laser surgery~\cite{Lopriore2007}. This may be explained considering the challenges in identifying anastomoses in conditions of poor visibility and constrained maneuverability of the fetsocope, especially in the presence of anterior placenta due to the unfavourable viewing angle. 
In this complex scenario, Computer-Assisted Intervention and Surgical Data Science methodologies may be exploited to provide surgeons with context awareness and decision support. However, the research in this field is still at its early stages and several challenges still have to be tackled~\cite{Pratt2015}.

To foster research in this field, we organized the \textbf{FetReg: Placental Vessel Segmentation and Registration in Fetoscopy}\footnote{FetReg2021 Challenge website: \url{https://fetreg2021.grand-challenge.org/}} inside the EndoVis MICCAI Grand Challenge\footnote{MICCAI EndoVis Challenge website: \url{https://endovis.grand-challenge.org/}}. This paper describes the FetReg Challenge in terms of tasks, datasets and evaluation procedures.

\section{Literature Review}
In this section, we provide a short overview of current literature in the field of placenta vessel segmentation and mosaicking. 

\subsection{Placenta Vessel Segmentation}
The Surgical Data Science~\cite{maier2017surgical} community is
working towards developing computer-assisted algorithms to perform intra-operative
tissue segmentation. However, Surgical Data Science approaches for fetal image segmentation, including inter-fetus (twin) membrane and placenta-vessel segmentation, have only been marginally explored. 
A number of approaches have been proposed for inter-twin membrane segmentation from in-vivo placenta~\cite{casella2020inter,casella2021shape}, but vessel segmentation is currently covering a larger space in the literature. 
Vessel segmentation is used both to directly provide guidance to the surgeon and as a prior to perform mosaicking for field-of-view expansion. Examples include~\cite{sadda2019deep,bano2019deep}.
In~\cite{sadda2019deep}, a shallow U-Net architecture is proposed to perform patch-based vessel segmentation from intra-operative fetoscopic frames. U-Net is also used in~\cite{bano2020deep}, where vessel probability maps are exploited to perform image mosaicking. 

Despite the promising results, limitations still exist which hamper the translation of the proposed methodology in the clinical practice. 
Besides the development of advanced algorithms, a major effort is needed to collect large, high-quality, multi-centre datasets, which are currently not available to the research community. Having such multi-centre datasets is crucial with a view to attenuate the well-known problem of covariance shift and develop robust and generalizable algorithms.

\subsection{Video Mosaicking}
Several Surgical Data Science approaches have been proposed for fetoscopic mosaicking in the years. 
A first distinction can be made between approaches that rely on image information only and those that exploit additional hardware, such as~\cite{tella2019pruning} that rely on electromagnetic tracking. 
Among methods that exploit only information from the fetoscope, some researchers have proposed indirect registration methods based on extraction and matching of image features~\cite{daga2016real,reeff2006mosaicing}. These approaches, however, have only been validated on synthetic phantom or ex-vivo placental sequences where the imaging resolution, visual quality and appearance are vastly different from in-vivo placental imaging. On the other hand, direct registration methods~\cite{peter2018retrieval} minimize photometric consistency between frames and have been more successful at dealing with in-vivo fetoscopy data. More recently, deep learning algorithms have been proposed, both for homography estimation~\cite{bano2019deep,bano2020deep} and detection of stable regions as a prior for frame registration~\cite{gaisser2018stable}. These approaches are validated on in-vivo~\cite{bano2019deep,bano2020deep} or an underwater phantom setting~\cite{gaisser2018stable}. In~\cite{bano2020deep}, it is shown that placental vessels provide unique landmarks which help in overcoming visibility challenges in in-vivo placental imaging. As a result, the use of segmented vessel maps for consecutive frame registration generate reliable mosaics. 

Robotics has shown the potential to improve stability of the imaging device by providing precise control of the fetoscopy instrument~\cite{dwyer2017continuum}. Efforts have also been made towards the design of a robotic multimodal endoscope that includes an optical ultrasound and white light stereo camera~\cite{dwyer2019robotic}. This endoscope has shown to provide improved visualization in an in-lab phantom experiment. Due to the limited form-factor of the clinically approved fetoscope, these solutions are not yet applicable in clinical settings.

While promising results have been achieved for mosaicking from short video sequences, long-term mapping still remains an open challenge. This is because of the intra- and inter-case variability in each procedure, dynamically challenging non-planar views, poor visibility, texture paucity, low resolution and occlusion due to the presence of fetus and ablation tool in the field-of-view.

\section{Challenge Tasks}
The \textit{FetReg challenge} aims to advance the current state-of-the-art in placental vessel segmentation and mosaicking~\cite{bano2020deep} by providing a benchmark multi-centre large-scale dataset that captures variability across different patients and different clinical institutions. The participants are required to complete two sub-tasks on the provided dataset:

\begin{itemize}
    \item \textbf{Task 1: Placental semantic segmentation}: The participants are required to segment four classes, namely, background, vessels, tool (ablation instrument) and fetus. This task will be evaluated on an unseen test data that is independent of the training data videos. The aim is to assess the generalization capability of the trained segmentation model on unseen fetoscopic video frames. The evaluation metric for the segmentation task is provided in Section~\ref{sec:seg_metric}.
    
    \item \textbf{Task 2: Registration for Mosaicking}: The participants are required to perform registration of consecutive frames to create an expanded field-of-view image of the fetoscopic environment. The task will be evaluated on unseen video clips extracted from fetoscopic procedure videos which are not part of the training data. The aim is to assess the robustness and performance of the participant's registration method for creating a drift-free mosaic from unseen data. Details of the evaluation metric for the registration task are provided in Section~\ref{sec:reg_metric}. 
\end{itemize}
  
\section{Dataset Collection}
The \textit{FetReg} dataset is unique as it is the first large-scale fetoscopic video dataset of 18 different procedures. The videos contained in this dataset are collected from three fetal surgery centres across Europe, namely,
\begin{itemize}
    \item Fetal Medicine Unit, University College London Hospital (UCLH), London, UK,
    \item Department of Fetal and Perinatal Medicine, Istituto "Giannina Gaslini", Genoa, Italy,
    \item Department of Development and Regeneration, University Hospital Leuven, Leuven, Belgium.
\end{itemize}
Alongside capturing the intra-case and inter-case variability, the multi-centre data collection allowed capturing the variability that arises due to different clinical settings and imaging equipment at different clinical sites. 
At UCLH, the data collection was carried out as part of the GIFT-SURG\footnote{GIFT-Surg project: \url{https://www.gift-surg.ac.uk/}} project. The requirement for formal ethical approval was waived as the data were fully anonymized in the corresponding clinical centres before being transferred to the organizers of the \textit{FetReg} challenge. The fully anonymized fetoscopic data used in this challenge will be published for research and educational purposes after the challenge\footnote{FetReg data is currently available only to the EndoVis2021 participants upon signing the EndoVis2021 rule agreement. The public data release is subject to the publication of the joint journal article on the challenge results and data analysis.}. 

\begin{table}[t!]
	\centering
	\caption{Summary of the challenge dataset. For each video, image resolution, number of annotated frames (for the segmentation task), occurrence of each class per frame and average number of pixels presence per class per frame are presented. For the registration task, number of unlabelled frames in each video clip are provided. Key: BG - background.}
	\label{tab:seg_dataset}
	\resizebox{1.0\textwidth}{!}{
	\footnotesize
	\begin{tabular}{|r|c|c|c| >{\raggedleft\arraybackslash}m{0.7cm}| >{\raggedleft\arraybackslash}m{0.7cm}| >{\raggedleft\arraybackslash}m{0.7cm}| >{\raggedleft\arraybackslash}m{1.0cm}| >{\raggedleft\arraybackslash}m{0.8cm}| >{\raggedleft\arraybackslash}m{0.7cm}| >{\raggedleft\arraybackslash}m{0.7cm}|c|}
	\hline
	\textbf{Sr.} &\textbf{Video} &\textbf{Image} &\textbf{No. of} &\multicolumn{3}{c|}{\textbf{Occurrence}} &\multicolumn{4}{c|}{\textbf{Occurrence}} &\textbf{Unlabel-} \\ 
	&\textbf{name} &\textbf{Resolution} &\textbf{Labelled} &\multicolumn{3}{c|}{\textbf{(frame)}}  &\multicolumn{4}{c|}{\textbf{(Avg. pixels)}} &\textbf{-led clips} \\ 
	\cline{5-11}
	& &\textbf{(pixels)} &\textbf{frames} &\textbf{Vessel} &\textbf{Tool} &\textbf{Fetus} &\multicolumn{1}{c|}{\textbf{BG}} &\textbf{Vessel} &\textbf{Tool} &\textbf{Fetus} &\textbf{\# frames}\\ \hline \hline
	
	1. &Video001 &$470 \times 470$ &152 &152	&21	&11	&196463 &21493 &1462 &1482  &346 \\ \hline
	2. &Video002 &$540 \times 540$ &153 &153 &35 &1 &271564 &16989 &3019 &27 &259 \\ \hline
	3. &Video003 &$550 \times 550$ &117 &117 &52 &32 &260909 &27962 &3912 &9716 &541 \\ \hline
	4. &Video004 &$480 \times 480$ &100 &100 &21 &18 &212542 &14988 &1063 &1806 &388 \\ \hline
	5. &Video005 &$500 \times 500$ &100 &100 &35 &30 &203372 &34350 &2244 &10034 &722 \\ \hline
	6. &Video006 &$450 \times 450$ &100 &100 &49 &4 &171684 &28384 &1779 &653 &452 \\ \hline
	7. &Video007 &$640 \times 640$ &140 &140 &30 &3 &366177 &37703 &4669 &1052 &316 \\ \hline
	8. &Video008 &$720 \times 720$ &110 &105 &80 &34 &465524 &28049 &13098 &11729 & 295 \\ \hline	
	9. &Video009 &$660 \times 660$ &105 &104 &40 &14 &353721 &68621 &7762 &5496 &265 \\ \hline
	10. &Video011 &$380 \times 380$ &100 &100	&7 &37 &128636 &8959 &184 &6621 &424 \\ \hline
	11. &Video013 &$680 \times 680$ &124 &124	&54	&21 &411713 &36907 &8085 &5695 &247 \\ \hline	
	12. &Video014 &$720 \times 720$ &110 &110 &54 &14 &464115 &42714 &6223 &5348 &469 \\ \hline
	13. &Video016 &$380 \times 380$ &100 &100	&16	&20	&129888 &11331 &448 &2734 &593 \\ \hline
	14. &Video017 &$400 \times 400$ &100 &97 &20 &3 &151143 &7625	&753 &479 &490 \\ \hline
	15. &Video018 &$400 \times 400$ &100 &100	&26	&11	&139530 &15935 &1503 &3032 &352 \\ \hline
	16. &Video019 &$720 \times 720$ &149 &149	&15	&31	&470209 &38513 &1676 &8002 &265 \\ \hline
	17. &Video022 &$400 \times 400$ &100 &100	&12	&1	&138097 &21000 &650 &253 &348 \\ \hline
	18. &Video023 &$320 \times 320$ &100 &92 &14 &8 &94942 &6256 &375 &828 &639 \\ \hline \hline
	\multicolumn{3}{|c|}{\textbf{All videos}} &\textbf{2060} &2043 &581 &293 &4630229 &467779 &58905 &74987 &\textbf{7411} \\ \hline 
	\end{tabular}
	}
\end{table}

\subsection{Segmentation Dataset Description}
\label{sec:dataset_description}
Fetoscopy videos acquired from the three different fetal medicine centres are first decomposed into frames and excess black background is cropped to obtain squared images capturing mainly the fetoscope field-of-view. From each video a subset of 100-150 non-overlapping informative frames was selected and manually annotated. All pixels in each image are labelled with background (0), placental vessel (1), ablation tool (2) or fetus class (3). There is no overlap between the segmentation labels, hence all labels are mutually exclusive.

Annotations were performed by four academic researchers and staff members with a background in fetoscopic imaging. Additionally, annotation services were obtained from Humans in the Loop (HITL)\footnote{Humans in the Loop: \url{https://humansintheloop.org/}} for a subset of videos. HITL is an award-winning social enterprise founded in 2017 who provides annotations using skilled workers displaced by conflict with the vision to connect the conflict-affected communities to digital work. All the annotations were further verified by two fetal medicine specialists who confirmed the correctness and consistency of the labels. The publicly available Supervisely\footnote{Supervisely: a web-based annotation platform: \url{https://supervise.ly/}} platform was used for annotating the dataset. 

\begin{figure}[hbtp]
\centering
\includegraphics[width=1.0\textwidth]{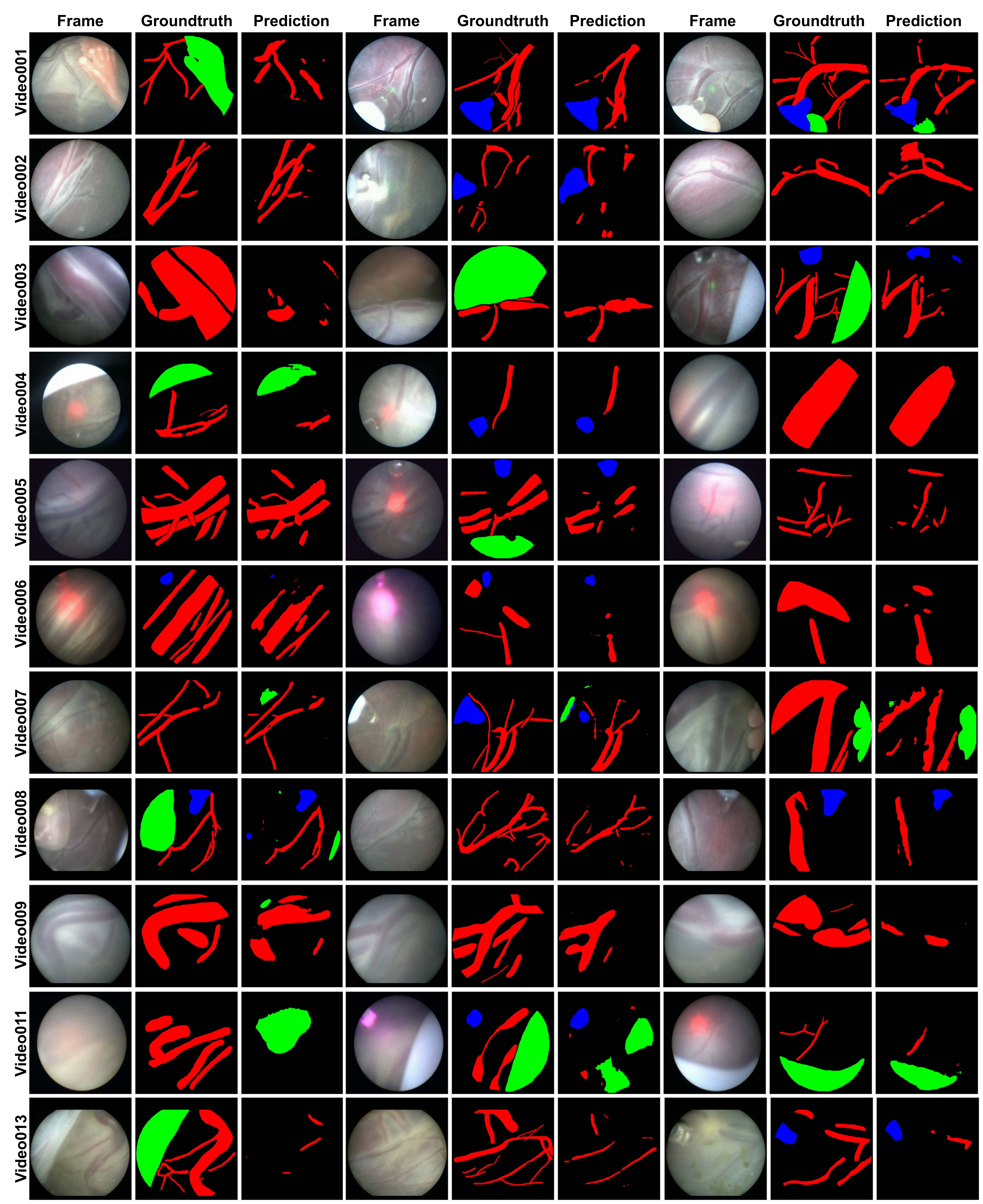}
\caption{Representative images along with the segmentation annotations (Groundtruth) and baseline segmentation output (Prediction) for Video001, 002, 003, 004, 005, 006, 007, 008 and 009 videos. Background (black), vessel (red), tool (blue) and fetus (green) labels are shown. Observe the intra- and inter-case variability in the videos.} 
\label{fig:seg_pred_1}
\end{figure}

\begin{figure}[hbtp]
\centering
\includegraphics[width=1.0\textwidth]{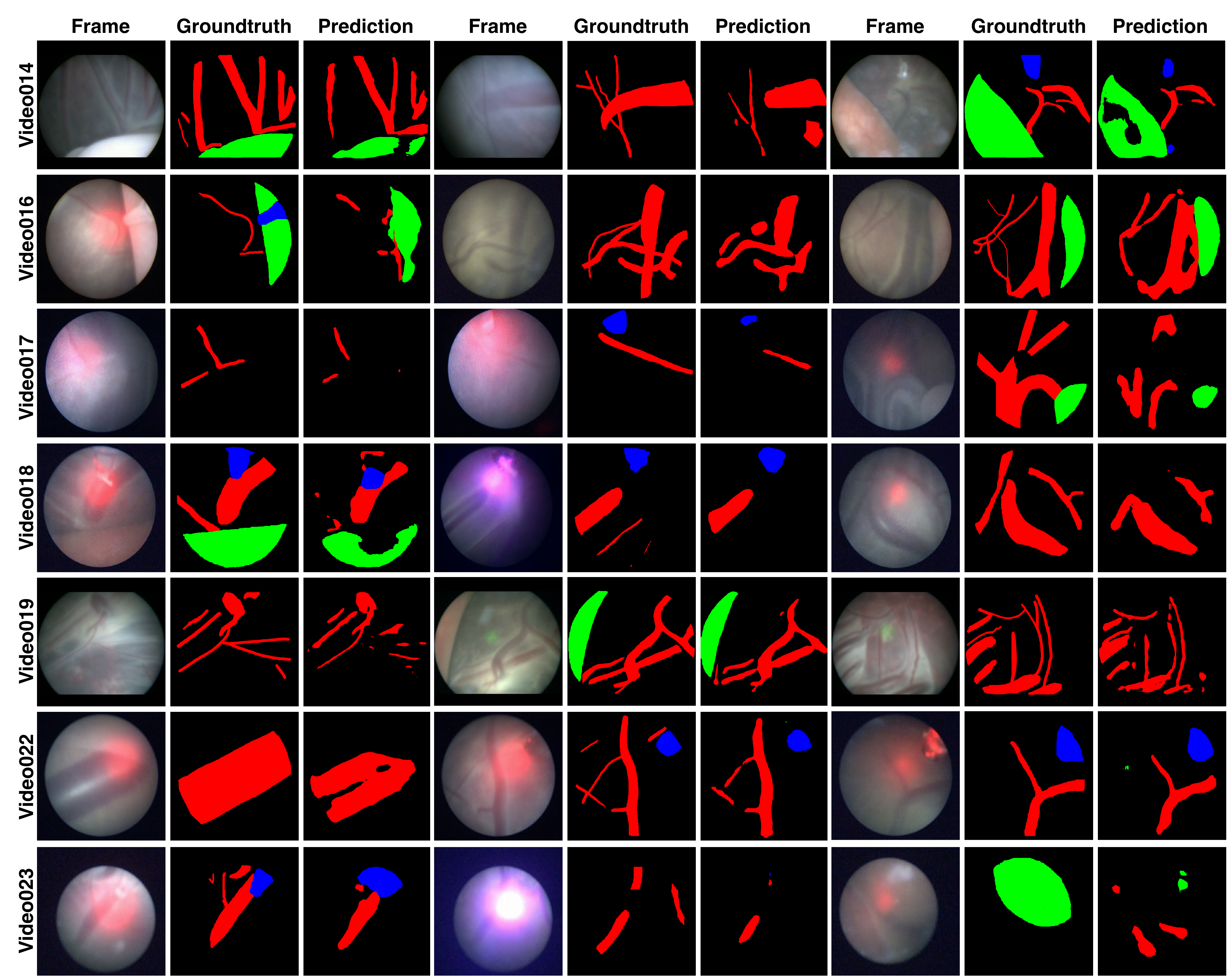}
\caption{Representative images along with the segmentation annotations (Groundtruth) and baseline segmentation output (Prediction) for Video014, 016, 017, 018, 019, 022 and 023 videos. Background (black), vessel (red), tool (blue) and fetus (green) labels are shown. Observe the intra- and inter-case variability in the videos.} 
\label{fig:seg_pred_2}
\end{figure}

\textit{FetReg} dataset for the segmentation task contains 2060 annotated images from 18 different in-vivo TTTS fetoscopic surgeries. Table~\ref{tab:seg_dataset} summarizes the segmentation dataset. Note that the frames present different resolutions as the fetoscopic videos are captured at different centres with different facilities (e.g., device, light scope). 
From the occurrences of the classes in Table~\ref{tab:seg_dataset}, it can be observed that the dataset is highly unbalanced, the \textit{Vessel} is the most frequent class while \textit{Tool} and \textit{Fetus} are presented only in a small subset of images corresponding to 28\% of the dataset.
Fig.~\ref{fig:seg_pred_1} and Fig.~\ref{fig:seg_pred_2} show some representative annotated frames images from each video. Note that the frame appearance and quality changes in each video due to the large variation in intra-operative environment among different cases. Amniotic fluid turbidity resulting in poor visibility, artefacts introduced due to spotlight light source and reddish reflection introduced by the laser tool, low resolution, texture paucity, non-planar views due to anterior placenta imaging, are some of the major factors that contribute to increase the variability in the data. Large intra-case variation can also be observed from these representative images. All these factors contribute towards limiting the performance of the existing placental image segmentation and registration methods~\cite{bano2020deep,bano2019deep,bano2020deepijcars}. 

The \textit{FetReg} challenge provides an opportunity to make advancements in the current literature by designing and contributing novel segmentation and registration methods that are robust even in the presence of the above-mentioned challenges. 


\begin{figure}[t]
\centering
\includegraphics[width=1.0\textwidth]{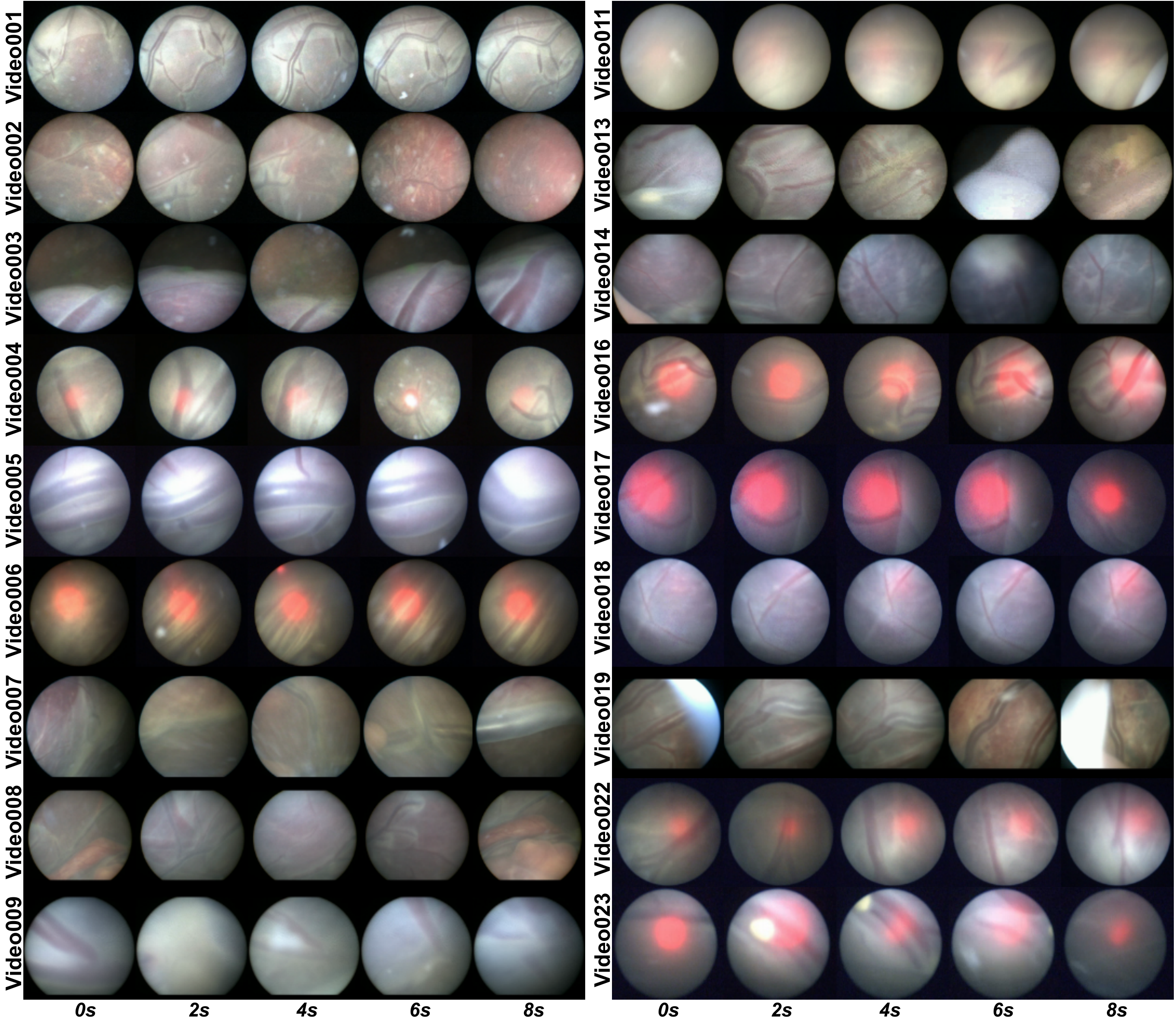}
\caption{Representative frames at every 2 \textit{seconds} from the 18 video clips. These clips are unannotated and are released for the registration and mosaicking task. Number of frames in each clip are mentioned in Table~\ref{tab:seg_dataset}. } 
\label{fig:seq_representation}
\end{figure}

\subsection{Registration Dataset Description}
Average duration of the TTTS fetoscopic surgery is approximately 30 minutes. Not all fetoscopic frames are suitable for frame registration and mosaicking. This is because of the presence of fetuses, laser ablation fibre and working channel port which can occlude the field-of-view of the fetoscope. Frame registration and mosaicking is only required in occlusion-free video segments that capture the surface of the placenta~\cite{bano2020fetnet} as these are the segments in which the surgeon is exploring the intraoperative environment to identify abnormal vascular connections. Expanding the field-of-view through mosaicking in these video segments can facilitate the procedure by providing better visualization of the environment. 

For the registration and mosaicking task, we have provided one video clip per video for all 18 procedures in our dataset. These frames are not annotated with segmentation labels. The number of frames in each video clip is reported in Table~\ref{tab:seg_dataset}. Representative frames at every 2 seconds from the 18 video clips are shown in Figure~\ref{fig:seq_representation}. Observe the variability in the appearance, lighting conditions and image quality in all video clips. Even though there is no noticeable deformation in fetoscopic videos, which is usually thought to occur due to breathing motion, the views can be non-planar as the placenta can be anterior or posterior. Moreover, there is no groundtruth camera motion and scene geometry that can be used to evaluate video registration approaches for in-vivo fetoscopy. In Section~\ref{sec:reg_metric}, we detail how this challenge is addressed with an evaluation metric that is correlated with good quality, consistent, and complete mosaics~\cite{bano2020deep}.  

\section{Segmentation}

Segmenting intraoperative fetoscopy images mainly involves blood vessels, fetuses (and related fetal parts) and surgical (laser) tools.
Segmenting these structures can support indirect registration methods to accomplish the mosaicking task. Hence, the research community has been working to develop algorithms for robust and accurate segmentation, with a particular focus on vessel segmentation.

Obtaining an accurate segmentation is challenging due to homogeneous texture of the tissues and reduced visibility within the amniotic sac (small field-of-view of the fetoscope, turbidity of the amniotic fluid). 
The quality of fetoscopic images is further degraded by the presence of occlusions (laser fibre, fetuses and particles) and light reflections. In anterior placental procedures, where the $30^{\circ}$ fetoscope is used, the field-of-view is further reduced due to the view angle between the camera and the placenta surface.

In \textit{FetReg}, placental semantic segmentation is treated as a multi-class problem (refer to Sec.~\ref{sec:dataset_description}).

\subsection{Segmentation metric}
\label{sec:seg_metric}
For evaluating the performance of segmentation models, we compute for each frame provided in the test set the Mean Intersection over Union ($\mathit{IoU}$) per class between the prediction and the manually annotated groundtruth. 

\begin{equation}
    \mathit{IoU}
    =\frac{\mathit{TP}}{(\mathit{TP}+\mathit{FP}+\mathit{FN})}
\end{equation}
where $\mathit{TP}$ are the correctly classified pixels belonging to a class, $\mathit{FP}$ are the pixels incorrectly predicted as in that specific class and $\mathit{FN}$ are the pixels in that class incorrectly classified as not belonging to it.

\subsection{Baseline method}
As the baseline model for analysing the \textit{FetReg} segmentation sub-task, we trained a U-Net~\cite{ronneberger2015u} with ResNet50~\cite{he2016deep} backbone. U-Net was selected as the baseline as it is the most commonly used network for biomedical imaging segmentation. 
Softmax activation was used at the final layer of the network (considering the multi-class nature of the segmentation dataset).
Cross entropy loss was computed and backpropagated during training.  

\subsection{Experimental Setup}
\label{sec:exp_setup}
Our baseline model was trained for 300 epochs on the dataset provided for Task 1. The performance for all patients is evaluated by training the model using the cross-fold validation scheme shown in Table~\ref{tab:seg_results}. We created 6 folds, where each fold contained 3 procedures, with the aim to preserve as much variability as possible for each fold while keeping the number of samples in each fold approximately balanced. 
During training, the images were resized to $448 \times 448$ pixels to reduce the probability of getting unannotated crops when compared to the original size, keeping negligible the loss of information introduced by resampling.
To perform data augmentation, at each iteration step, a patch of $256 \times 256$ pixels was extracted at a random position in the image. Each of the extracted patches was augmented by applying a random rotation in range $(-45^{\circ} , +45^{\circ})$, horizontal and vertical flip, scaling with a factor in the range of $(-20\%, +20\%)$ and random variation in brightness $(-20\%, +20\%)$ and contrast $(-10\%, +10\%)$. 
Segmentation results on the dataset released for Task 1 were obtained by patch-wise inference using $256 \times 256$ pixels patches with stride$=$8. 

\begin{table}[]
\centering
\caption{Results of segmentation for the Task 1 dataset. Mean $\mathit{IoU}$ for each class over each video and, in the last row, the average mean $\mathit{IoU}$ per class are reported. Key: BG-background.}
\label{tab:seg_results}
\resizebox{1.0\textwidth}{!}{
\begin{tabular}{|c|c|c|c|c|c|c|c|c|c|c|c|c|}
\hline
\multirow{2}{*}{\textbf{Video}} & \multicolumn{4}{c|}{\textbf{Class}} & \textbf{Overall} & \multirow{2}{*}{\textbf{Fold}} & \textbf{Images} & \multicolumn{4}{c|}{\textbf{Class}} & \textbf{Overall} \\ \cline{2-5} \cline{9-12}
 & \textbf{BG} & \textbf{Vessel} & \textbf{Tool} & \textbf{Fetus} & \textbf{per video} &  & \textbf{per fold} & \textbf{BG} & \textbf{Vessel} & \textbf{Tool} & \textbf{Fetus} & \textbf{per fold} \\ \hline
Video001 & 0.83 & 0.85 & 0.69 & 0.74 & 0.64 & \multirow{3}{*}{1} & \multirow{3}{*}{352} & \multirow{3}{*}{0.80} & \multirow{3}{*}{0.83} & \multirow{3}{*}{0.64} & \multirow{3}{*}{0.74} & \multirow{3}{*}{0.61} \\ \cline{1-6}
Video006 & 0.67 & 0.67 & 0.74 & 0.76 & 0.58 &  &  &  &  &  &  &  \\ \cline{1-6}
Video016 & 0.80 & 0.83 & 0.64 & 0.74 & 0.60 &  &  &  &  &  &  &  \\ \hline
Video002 & 0.78 & 0.79 & 0.80 & 0.53 & 0.56 & \multirow{3}{*}{2} & \multirow{3}{*}{353} & \multirow{3}{*}{0.80} & \multirow{3}{*}{0.81} & \multirow{3}{*}{0.83} & \multirow{3}{*}{0.78} & \multirow{3}{*}{0.69} \\ \cline{1-6}
Video011 & 0.75 & 0.72 & 0.73 & 0.83 & 0.64 &  &  &  &  &  &  &  \\ \cline{1-6}
Video018 & 0.80 & 0.81 & 0.83 & 0.78 & 0.71 &  &  &  &  &  &  &  \\ \hline
Video004 & 0.80 & 0.80 & 0.72 & 0.80 & 0.66 & \multirow{3}{*}{3} & \multirow{3}{*}{349} & \multirow{3}{*}{0.76} & \multirow{3}{*}{0.78} & \multirow{3}{*}{0.79} & \multirow{3}{*}{0.55} & \multirow{3}{*}{0.65} \\ \cline{1-6}
Video019 & 0.81 & 0.81 & 0.64 & 0.85 & 0.65 &  &  &  &  &  &  &  \\ \cline{1-6}
Video023 & 0.76 & 0.78 & 0.79 & 0.55 & 0.56 &  &  &  &  &  &  &  \\ \hline
Video003 & 0.79 & 0.81 & 0.72 & 0.79 & 0.66 & \multirow{3}{*}{4} & \multirow{3}{*}{327} & \multirow{3}{*}{0.82} & \multirow{3}{*}{0.82} & \multirow{3}{*}{0.80} & \multirow{3}{*}{0.93} & \multirow{3}{*}{0.66} \\ \cline{1-6}
Video005 & 0.71 & 0.77 & 0.79 & 0.56 & 0.56 &  &  &  &  &  &  &  \\ \cline{1-6}
Video014 & 0.82 & 0.82 & 0.80 & 0.93 & 0.78 &  &  &  &  &  &  &  \\ \hline
Video007 & 0.78 & 0.77 & 0.84 & 0.72 & 0.66 & \multirow{3}{*}{5} & \multirow{3}{*}{350} & \multirow{3}{*}{0.78} & \multirow{3}{*}{0.81} & \multirow{3}{*}{0.85} & \multirow{3}{*}{0.54} & \multirow{3}{*}{0.67} \\ \cline{1-6}
Video008 & 0.78 & 0.76 & 0.75 & 0.85 & 0.68 &  &  &  &  &  &  &  \\ \cline{1-6}
Video022 & 0.78 & 0.81 & 0.85 & 0.54 & 0.60 &  &  &  &  &  &  &  \\ \hline
Video009 & 0.80 & 0.80 & 0.80 & 0.73 & 0.66 & \multirow{3}{*}{6} & \multirow{3}{*}{329} & \multirow{3}{*}{0.66} & \multirow{3}{*}{0.66} & \multirow{3}{*}{0.73} & \multirow{3}{*}{0.57} & \multirow{3}{*}{0.58} \\ \cline{1-6}
Video013 & 0.72 & 0.77 & 0.75 & 0.50 & 0.50 &  &  &  &  &  &  &  \\ \cline{1-6}
Video017 & 0.66 & 0.66 & 0.73 & 0.57 & 0.48 &  &  &  &  &  &  &  \\ \hline
\textbf{per class} & 0.78 & 0.79 & 0.76 & 0.75 & \multicolumn{8}{c|}{} \\ \hline
\end{tabular}
}
\end{table}

\subsection{Results and Discussion}
Table~\ref{tab:seg_results} shows the cross-validation results over each fold and individual videos reporting the per-class and overall mIOU values. Predicted segmentation masks for some representative images for each video are shown in Fig.~\ref{fig:seg_pred_1} and Fig.~\ref{fig:seg_pred_2}.

For vessel segmentation, U-Net with ResNet50 backbone achieved a Mean $\mathit{IoU}$ $0.7892$. From Fig.~\ref{fig:seg_pred_1} and Fig.~\ref{fig:seg_pred_2}, it can be observed that overall vessel segmentation gave promising results. In challenging cases, such as when the laser glow was extremely strong (Video023), the vessels were not segmented properly. Another issue was found in the presence of vessels with different morphology and contrast with respect to the training set (e.g., Video003) that led to inaccurate vessel segmentation.

The class imbalance of the dataset, which was discussed in Sec.\ref{sec:dataset_description}, impacted on identification of the tool (Mean $\mathit{IoU}$ $0.7637$) and the fetus (Mean $\mathit{IoU}$ $0.7522$). As shown in Video001, Video003, Video005, Video008 and Video023, the fetus in the scene was not identified at all. It can be clearly observed that the different shades on the fetus causes holes in the segmentation masks as in Video011, Video014 and Video018. 
The tool was correctly identified, even if not accurately segmented, in all videos. This is explainable considering the regular structure of tool.

With the \textit{FetReg} challenge, we aim to dramatically improve these performances, fostering breakthroughs in deep learning to provide context awareness to fetoscopy surgeons.

\section{Registration and Mosaicking}
\subsection{Problem Formulation}
In its most general form, the registration of fetoscopic placental images is a complex non-linear mapping that involves not only dealing with changes in camera perspective but also lens distortion, fluid light refraction, irregular placental shape, and outlier occlusions (laser tool, fetus, floating particles). To further the problem complexity, calibrating the fetoscopic camera parameters before an in-vivo procedure is unfeasible not only due to the strict workflow in the operating room but also because some camera parameters change dynamically when focus is adjusted or the lens physically rotates.

Throughout the fetoscopic registration literature, most of these complexities have been ignored in favour of simple model approximations that render the problem tractable. Typically, a linear mapping between 2D homogeneous image coordinates is considered (i.e. a homography). This assumes absence of lens distortion and light refraction, and that the visualized scene is either fully planar or locally-planar. For algorithm stability and convergence, it is common to further constrain the problem to an affine registration (6 parameters), which is a sub-set of a general homography (8 parameters). In the context of this challenge we define the registration between fetoscopic frames to be of a general homography mapping, although further constraints or simplifications can be considered by participants as they see fit.

We assume that a registration algorithm receives as input a set of $N \in \{2,3,...,N\}$ consecutive image frames from a video $\{\s I_1,\s I_2,...\s I_N\}$ and outputs a set of $N-1$ incremental (pairwise) homography transformations $\{\m H_1,\m H_2,...,\m H_{N-1}\}$. Each homography $\m H_i$ is a $3\times3$ matrix that linearly maps the homogeneous coordinates of a 2D point $\v p_i = \tr{\begin{pmatrix}x_i & y_i & 1\end{pmatrix}}$ in image $\s I_i$ to its corresponding point $\v p_{i+1} = \tr{\begin{pmatrix}x_{i+1} & y_{i+1} & 1\end{pmatrix}}$ in image $\s I_{i+1}$:
\begin{equation}
\v p_{i+1} \sim \m H_{i} \v p_{i}
\end{equation}
where $\sim$ denotes equality up to a scale factor. Mapping points beyond consecutive frames is achieved with chained homography multiplication
\begin{equation}
\v p_{i+n} \sim \m{H}_{i \rightarrow i+n} \v p_{i}, \qquad \m{H}_{i \rightarrow i+n} \sim \m H_{i+n} ... \m H_{i+2} \m H_{i+1} \m H_{i}
\label{eq:Hchain}
\end{equation}

We can also warp an entire image $\s I_{i}$ so that it is aligned with an image $\s I_{i+n}$ by re-mapping and interpolating every pixel
\begin{equation}
\s I_{i \rightarrow i+n}=w(\s I_{i},\; \m{H}_{i \rightarrow i+n})
\end{equation}
There are multiple ways to define the warping function $w(\s I_{i},\; \m{H}_{i \rightarrow i+n})$, depending on the interpolation method. In this paper, we assume bilinear interpolation, which is the default option of the  \href{https://docs.opencv.org/master/da/d54/group__imgproc__transform.html#gaf73673a7e8e18ec6963e3774e6a94b87}{OpenCV function \texttt{cv2.warpPerspective()}}.

\subsection{Registration metric}
\label{sec:reg_metric}
Given $N$ consecutive frames and a set of $N-1$ homographies $\{\m H_1,\m H_2,...,\m H_{N-1}\}$, we can evaluate their consistency. We use the term consistency rather than accuracy since in the context of this challenge there are no groundtruth homographies. The ultimate clinical goal of fetoscopic registration is to generate consistent, comprehensible and complete mosaics that map the placental surface and guide the surgeon. We therefore consider for evaluation the registration consistency between pairs of non-consecutive frames that have a large overlap in field of view and present a clear view of the placental surface. The list of overlapping frame pairs on the test data can thus be considered as the \textit{groundtruth} information that is not disclosed to participants until challenge completion.

The overwhelming majority of overlapping frame pairs will be temporally close to each other, but can also include distant frames that revisit the same part of the scene. For algorithm evaluation, these frame pairs will be selected so that they always represent a clear view of the placental surface (no heavy occlusions) and their field of view overlapping area is always larger than $25\%$. In this paper, we will present consistency results for all image pairs that are 5 frames apart in a video (very large overlap), replicating the evaluation metric in~\cite{bano2020deep}.

Consider a source image $\s I_{i}$, a target image $\s I_{i+n}$, and a homography transformation $\m{H}_{i \rightarrow i+n}$ between them. We define the consistency between these two images as
\begin{equation}
    s_{i \rightarrow i+n} =\mathrm{sim}\left( w(\tilde{\s I}_{i},\m{H}_{i \rightarrow i+n}),\tilde{\s I}_{i+n}\right) 
\end{equation}
where $\mathrm{sim}()$ is an image similarity metric that is computed based on target image and warped source image, and $\tilde{\s I}$ is a smoothed version of the image $\s I$. We will define both these operations next:
\begin{itemize}
    \item $\textbf{Smoothing: } \tilde{\s I}$ is obtained by applying a $9\times9$ Gaussian filter with standard deviation of 2 to the original image $\s I$. This is fundamental to make the similarity metric robust to small outlier (e.g. particles) and image discretization artefacts. 
    \item \textbf{Similarity: } We start by determining the overlap region between the target $\tilde{\s I}$ and the warped source $w(\tilde{\s I}_{i},\m{H}_{i \rightarrow i+n})$, taking into account their circular edges. If the overlap contains less than $25\%$ of $\tilde{\s I}$, we consider that the registration failed as there will be no such cases in the evaluation pool. A rectangular crop is fit to the overlap, and the structural similarity index metric (SSIM) is calculated between the image pairs after having been smoothed, warped, and cropped. 
\end{itemize}

\subsection{Baseline method}
In this paper, we report the results obtained from the vessel map registration and mosaicking method reported in~\cite{bano2020deep} when applied to the unannotated 18 video clips provided as training set for Task 2. Vessel segmentation maps from consecutive frames are aligned by applying pyramidal Lucas-Kanade registration based direct registration approach. This approach minimizes the photometric loss between a fixed and a warped moving image. Since fetoscopic images have a circular field-of-view, the registration is performed using a circular mask that allows only analysing the flow field within the fetoscopic image field-of-view. Sequential registrations are then blended using the chain rule (eq.~\ref{eq:Hchain}) into a single large field-of-view image. 

\begin{figure}
    \centering
    \includegraphics[width=0.85\textwidth]{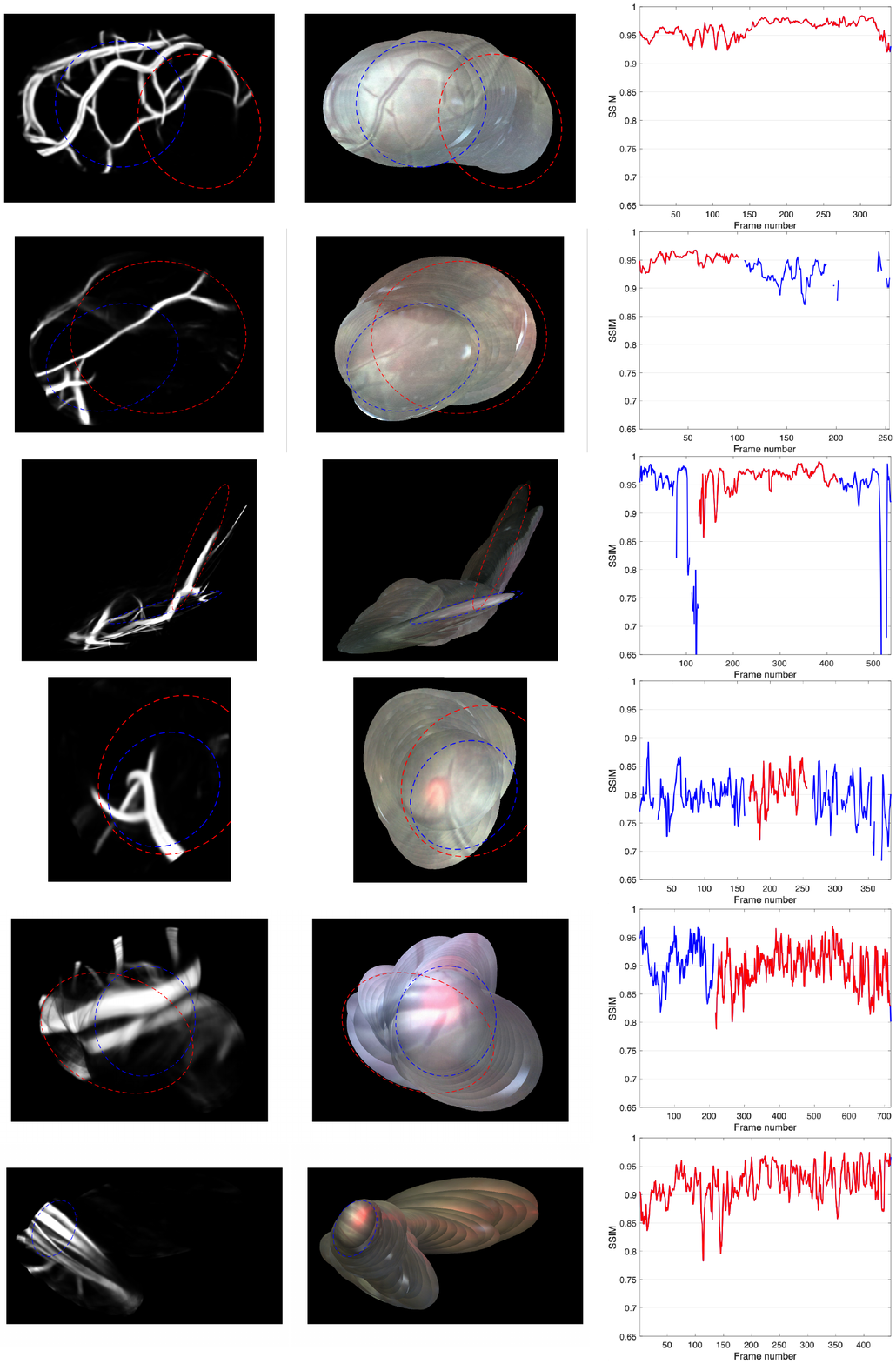}
    \caption{Left and centre show mosaics of vessel segmentations and RGB images respectively. Right shows the evaluation metric for every pair of images 5 frames apart in a sequence. Red represents the portion of frames visualized in the mosaics, while blue represents the remaining frames not shown in the mosaics. From top to bottom: Video001, Video002, Video003, Video004, Video005, Video006.}
    \label{fig:reg_eval_1}
\end{figure}
\begin{figure}
    \centering
    \includegraphics[width=0.85\textwidth]{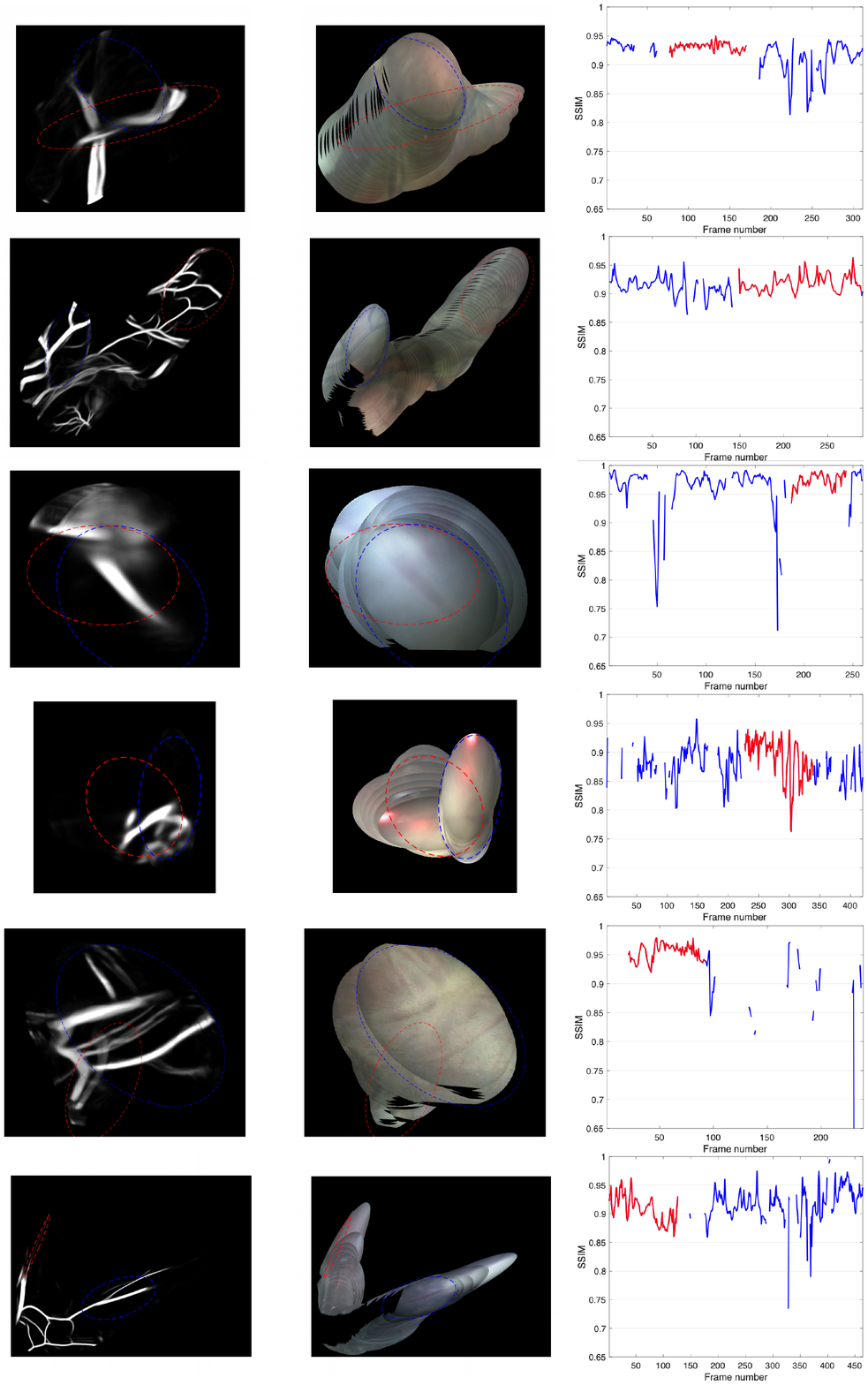}
    \caption{Left and centre show mosaics of vessel segmentations and RGB images respectively. Right shows the evaluation metric for every pair of images 5 frames apart in a sequence. Red represents the portion of frames visualized in the mosaics, while blue represents the remaining frames not shown in the mosaics. From top to bottom: Video007, Video008, Video009, Video011, Video013, Video014.}
    \label{fig:reg_eval_2}
\end{figure}
\begin{figure}
    \centering
    \includegraphics[width=0.85\textwidth]{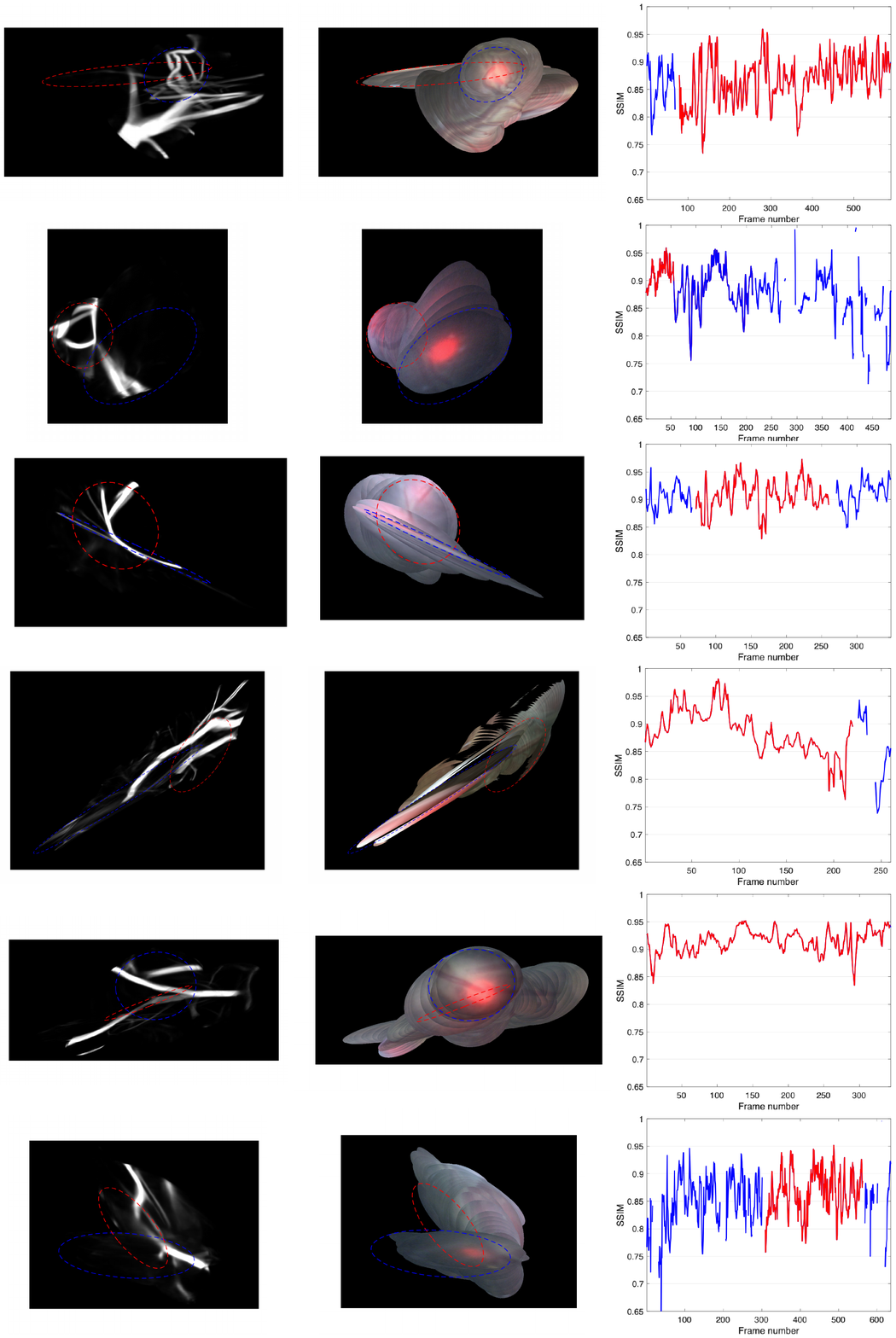}
    \caption{Left and centre show mosaics of vessel segmentations and RGB images respectively. Right shows the evaluation metric for every pair of images 5 frames apart in a sequence. Red represents the portion of frames visualized in the mosaics, while blue represents the remaining frames not shown in the mosaics. From top to bottom: Video016, Video017, Video018, Video019, Video022, Video023.}
    \label{fig:reg_eval_3}
\end{figure}

\subsection{Results and Discussion}
Qualitative and quantitative results for mosaicking are shown in \Cref{fig:reg_eval_1,fig:reg_eval_2,fig:reg_eval_3}. The mosaics of the vessel maps, RGB images and the evaluation metric plot for very pair of images 5 frames apart in a sequence are shown for all 18 video clips. For better visualization, mosaics are only displayed for the frames marked in red in the evaluation plots. For all 18 unannotated unseen video clips, the vessel maps are obtained through 6-fold cross-validation (presented in Sec.~\ref{sec:exp_setup}). It can be observed from these results that for sequences with clearly visible vessels (e.g. Video001, Video002, Video007, Video008), the obtained mosaics are consistent with a high evaluation score in the visualized video segment. There are some clips containing either heavy occlusions (Video008, Video013, Video019) or highly skewed views of the placenta (e.g. Video003) which can be observed in Fig.~\ref{fig:seq_representation}). Moreover, in frames with lack of texture and no visible vessels, our registration approach becomes invalid since it entirely relies on vessel segmentations. These issues cause registration to occasionally fail and therefore our baseline is often unable to reconstruct a continuous, smooth mosaic containing the entirety of a clip. 

While vessel-based registration facilitated in dramatically improving results over other existing approaches and helped in overcoming some visibility related challenges, fully robust mosaicking in fetoscopic videos still remains an open challenge. 
With the \textit{FetReg} challenge, we aim to open new frontiers in designing generalized models for fetoscopic videos that can overcome most of the associated challenges and can create consistent and drift-free mosaic for longer duration video clips.  

\begin{acknowledgements}
This work was supported by the Wellcome/EPSRC Centre for Interventional and Surgical Sciences (WEISS) at UCL (203145Z/16/Z), EPSRC (EP/P027938/1, EP/R004080/1,NS/A000027/1), the H2020 FET (GA 863146) and Wellcome [WT101957]. Danail Stoyanov is supported by a Royal Academy of Engineering Chair in Emerging Technologies (CiET1819/2/36) and an EPSRC Early Career Research Fellowship (EP/P012841/1). 
\end{acknowledgements}

\section{Conclusion}

In this paper, we presented a large multi-centre fetoscopic dataset extracted from 18 different TTTS laser therapy procedures, which is being proposed for the FetReg Challenge. While there has been recent progress in developing computer-assisted guidance algorithms for this procedure using segmentation and mosaicking techniques, none of them has been tested on such a large scale dataset to date. Results from state-of-the-art techniques for both vessel segmentation and sequential frame registration~\cite{bano2020deep} on this dataset and show that there are still many open challenges to be addressed. The very high variability in video appearance between different surgeries becomes one of the major elements to consider while designing new approaches in this domain. This sets the motivation for organizing the FetReg challenge and push the boundaries of the current state-of-the-art.

\newpage
\bibliographystyle{unsrt}  \bibliography{references.bib}  

\end{document}